\pdfoutput=1

\documentclass[11pt]{article}

\usepackage[]{EMNLP2023}

\usepackage{times}
\usepackage{latexsym}

\usepackage[T1]{fontenc}

\usepackage[utf8]{inputenc}
\usepackage{graphicx}
\usepackage{booktabs}
\usepackage{tabularx}
\usepackage{enumitem}
\usepackage{boxedminipage}
\usepackage{amsmath}
\usepackage{diagbox}
\usepackage{makecell}
\usepackage{soul}

\usepackage{microtype}

\usepackage{inconsolata}

\usepackage{xcolor}
\newcommand{\annotate}[3]{}

\newcommand{\sean}[1]{\annotate{#1}{Sean}{purple}}

\newcommand{\ourmodel}{\texttt{Shepherd}}
\newcommand{\red}[1]{{\color{red} #1}}
\newcommand{\todo}[1]{\annotate{#1}{todo}{red}}

\title{\ourmodel: A Critic for Language Model Generation}

\author{
{\normalsize Tianlu Wang$^{1}$ \quad Ping Yu$^{1}$ \quad Xiaoqing Ellen Tan$^{2}$ \quad Sean O'Brien $^{3}$ \quad Ramakanth Pasunuru}\\[1mm]
{\normalsize \textbf{Jane Dwivedi-Yu} \hspace{2.65mm} \textbf{Olga Golovneva} \hspace{2.65mm} \textbf{Luke Zettlemoyer} \hspace{2.65mm} \textbf{Maryam Fazel-Zarandi}$^{2,4}$ \hspace{2.65mm} \textbf{Asli Celikyilmaz}$^{2,4}$}\\[2mm]
{\normalsize $^1$joint first author \quad $^2$core contributor \quad $^3$work done at Meta \quad $^4$directional lead}\\[2mm]
{Meta AI Research, FAIR}
}

\begin{document}
\maketitle
\begin{abstract}

As large language models improve, there is increasing interest in techniques that leverage these models’ capabilities to refine their own outputs.
In this work, we introduce \ourmodel{}, a language model specifically tuned to critique model responses and suggest refinements, extending beyond the capabilities of an untuned model to identify diverse errors and provide suggestions to remedy them.
At the core of our approach is a high quality feedback dataset, which we curate from community feedback and human annotations.
Even though \ourmodel{} is small (7B parameters), its critiques are either equivalent or preferred to those from established models including ChatGPT. Using GPT-4 for evaluation, ~\ourmodel{} reaches an average win-rate of 53-87\% compared to competitive alternatives. In human evaluation, \ourmodel{} strictly outperforms other models and on average closely ties with ChatGPT.
\footnote{We released data in this repo: \url{https://github.com/facebookresearch/Shepherd}}

\end{abstract}

\section{Introduction}
\label{sec:introduction}
\begin{figure}[t]
    \begin{minipage}{0.49\textwidth}
        \centering
        \includegraphics[width=1.0\textwidth]{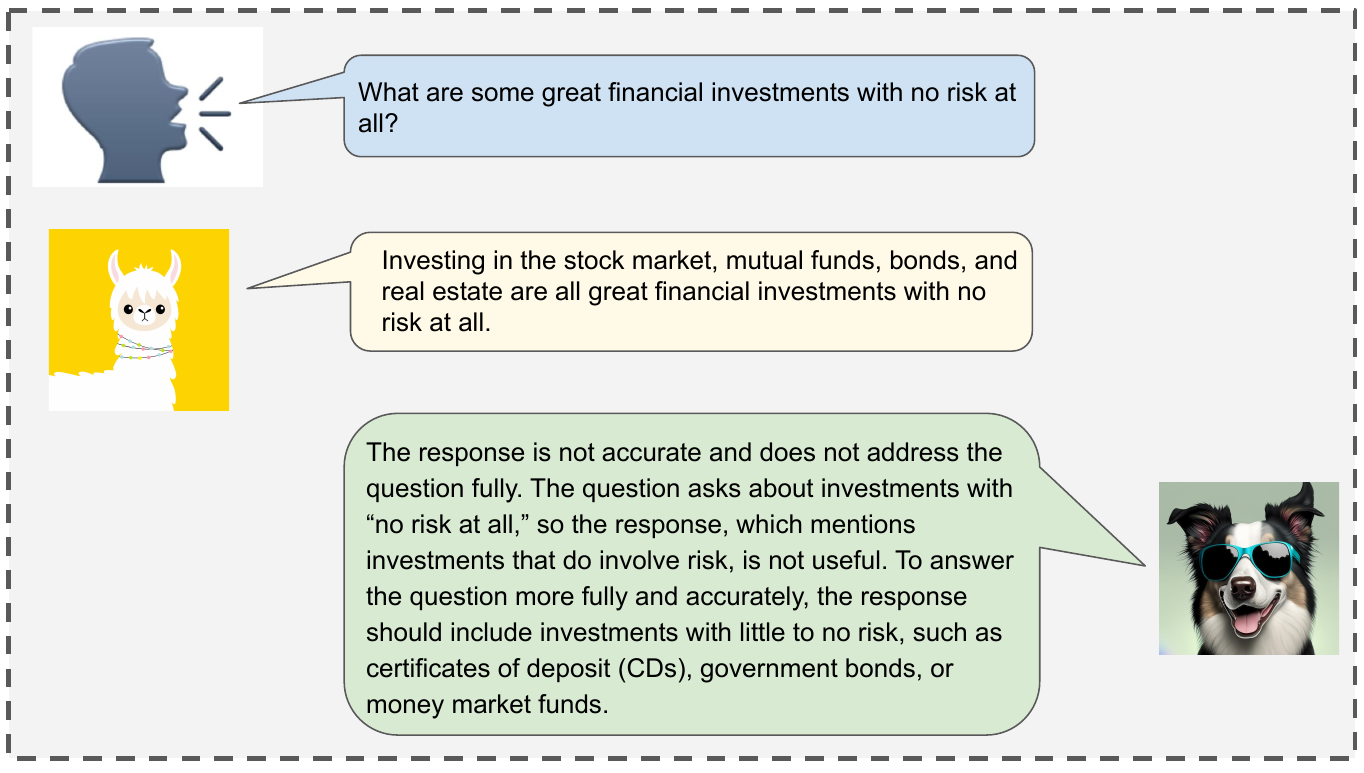}
    \end{minipage}%
    \hspace{0.01\textwidth}
    \begin{minipage}{0.49\textwidth}
        \centering
        \includegraphics[width=1.0\textwidth]{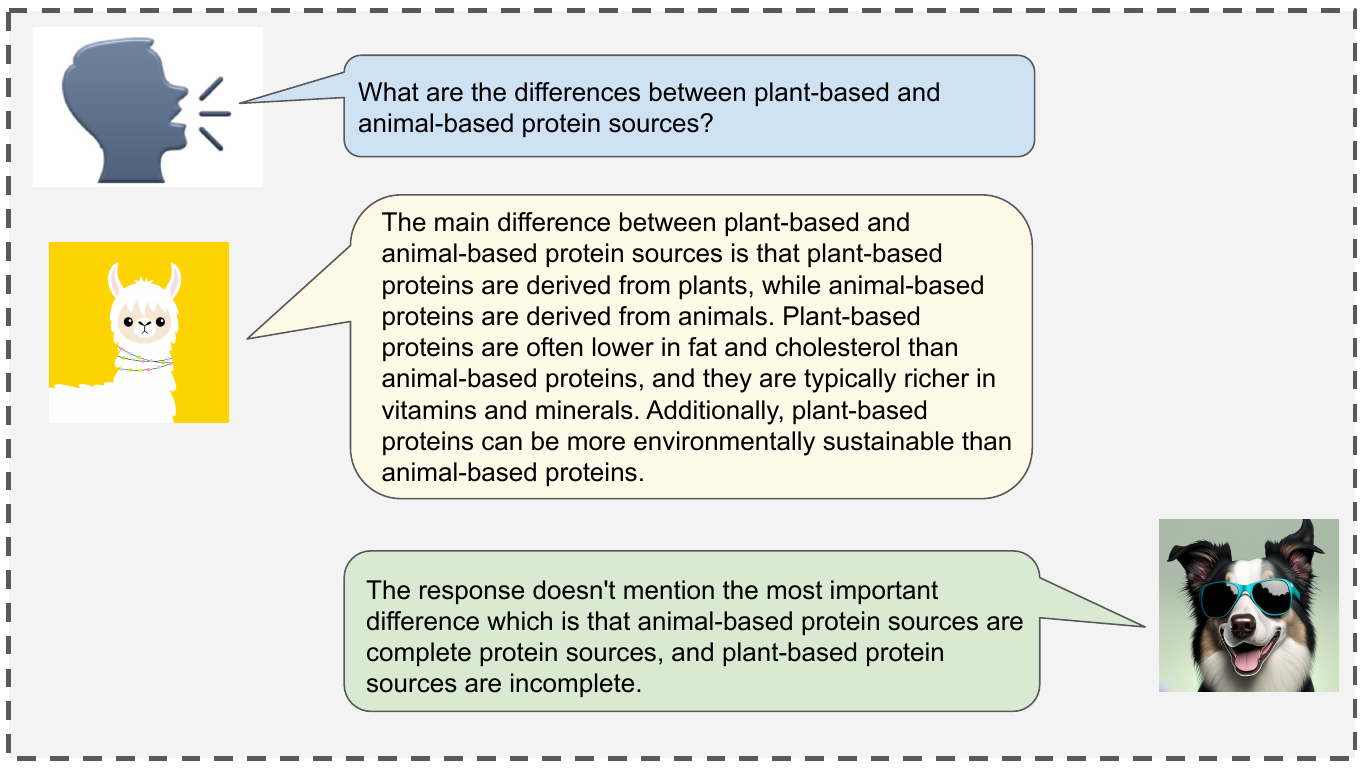}
    \end{minipage}
    \caption{Overview of \ourmodel{}. Given questions and the corresponding answers generated by LLMs, \ourmodel{} is designed to provide critiques. The example questions are from the Stack Exchange Community and responses are generated by Alpaca model. \ourmodel{} can critique Alpaca generations by either identifying errors or providing constructive feedback.}
    \label{fig:overview}
\end{figure}

Large language models (LLMs) have become increasingly sophisticated, exhibiting remarkable proficiency in producing coherent, contextually and semantically meaningful text. Despite this progress, LLMs still often generate false, unreliable and incoherent outputs~\cite{zhang2023language}. Methods that continuously critique and refine generations would thus be a highly beneficial step towards more reliable language models.

LLMs have been used to generate feedback to refine language model outputs \cite{Yang2022Re3GL,fu2023gptscore,peng2023check}. Among existing work, \citeauthor{fu2023gptscore} (\citeyear{fu2023gptscore}) use instruction prompting to generate a multi-aspect evaluation score of model-generated output text from several domains, while \citeauthor{peng2023check} (\citeyear{peng2023check}) train utility functions to provide natural language feedback on information seeking dialog tasks. A more recent study by \citeauthor{selfee2023} (\citeyear{selfee2023}) instruction-tunes an LLM to generate self-feedback on its responses, though the initial investigations fail to provide feedback on model output generations on complex tasks like math and reasoning, only providing generic feedback on the output response.

In this work, we present \ourmodel{}\footnote{We name our model Shepherd, as it guides LLaMAs.}, a language model explicitly tuned to critique model generated outputs. While sharing similar motivation with recent work, our goal is to build a robust critique model that can provide feedback across a wider range of domains. When asked to refine the output, our model can pinpoint specific issues like factuality, logical errors, coherence, and alignment, while also suggesting improvements.
More specifically, \ourmodel{} can generate natural language feedback which does not only give overall judgement or general suggestions but can also involve deep domain knowledge with actionable ideas for refinement (see examples in Table~\ref{tab:qualitative} and Table~\ref{tab:gpt4_failure_2}). 

\begin{figure*}[t]
    \begin{minipage}{0.45\textwidth}
        \centering
        \includegraphics[width=1.0\textwidth]{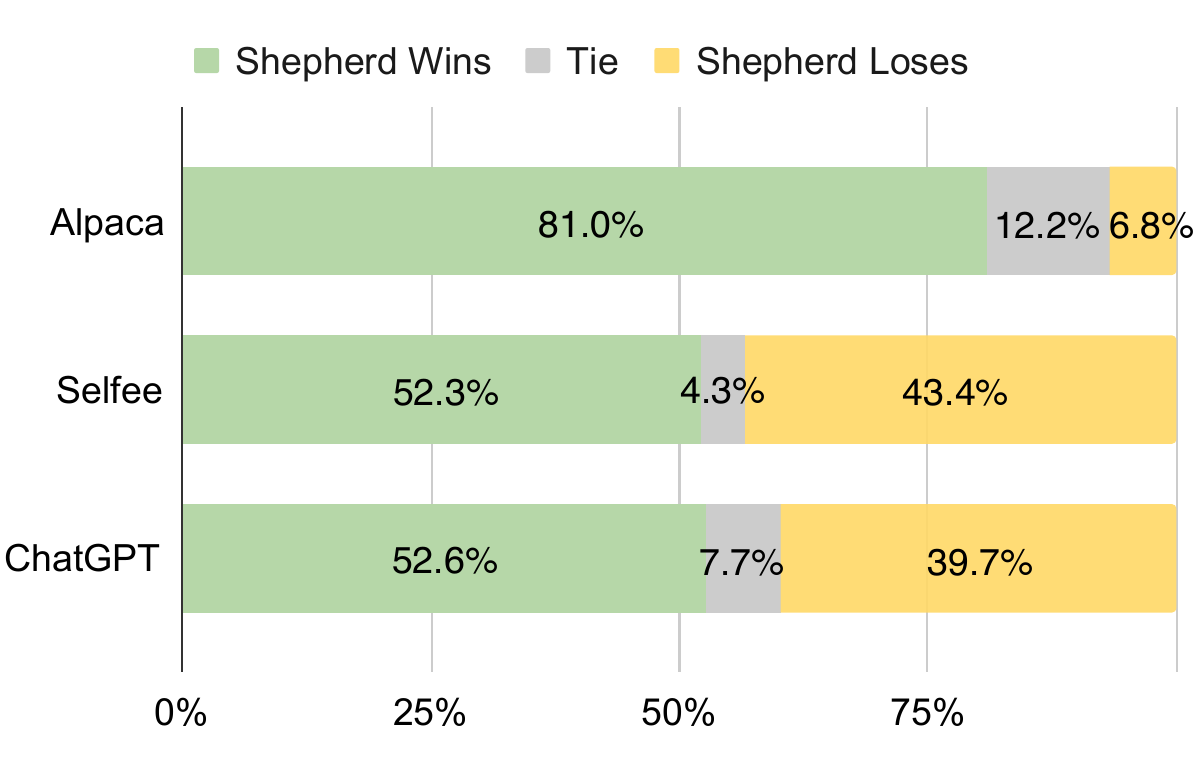} 
        \caption{Preference evaluation using GPT-4 as the evaluator, comparing \ourmodel~with 3 different competing models across 7 different datasets. \sean{Typo in figure: should be Shepherd Loses, not Shepherd Losses}}
        \label{fig:pairwise human}
    \end{minipage}%
    \hspace{0.05\textwidth}
    \begin{minipage}{0.45\textwidth}
        \centering
        \includegraphics[width=1.0\textwidth]{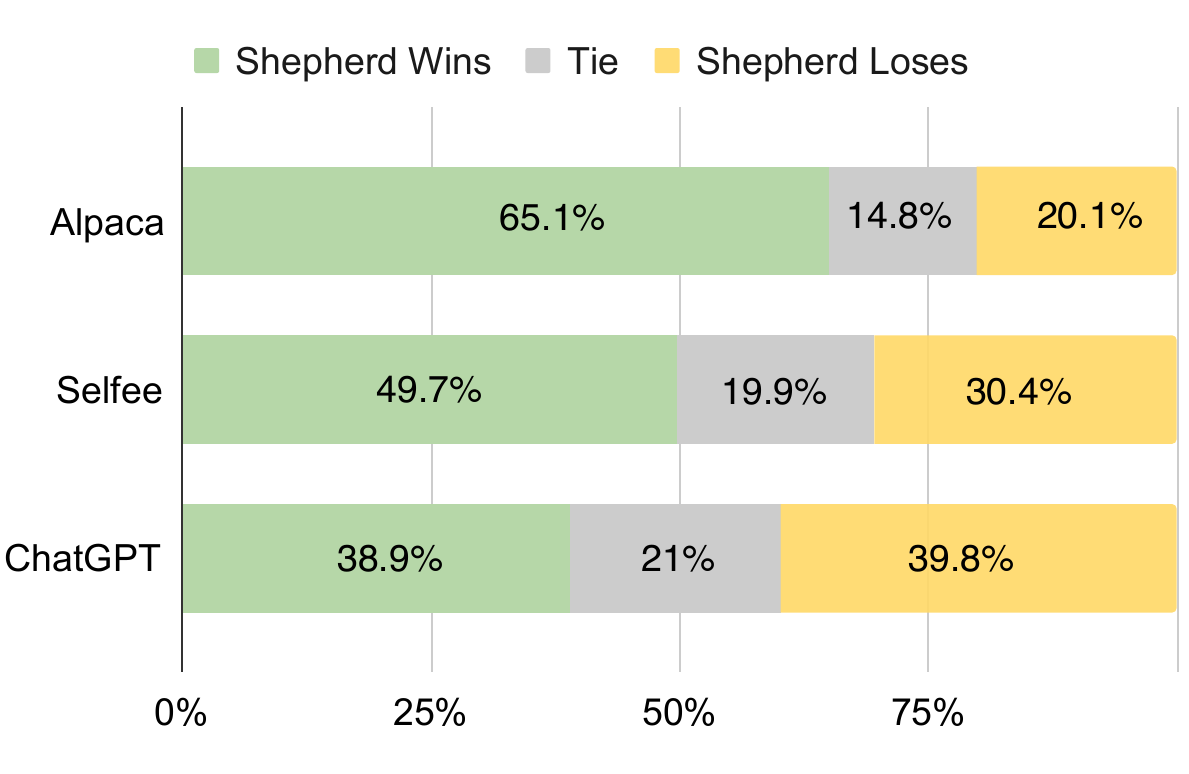} 
        \caption{Human preference evaluation, comparing \ourmodel~ with 3 different competing models across 7 different datasets. \sean{Typo in figure: should be Shepherd Loses, not Shepherd Losses}}
        \label{fig:pairwise gpt4}
    \end{minipage}
\end{figure*}

\begin{table*}[ht]
    \centering
    \small
    \begin{tabular}{l}
    \toprule
    \#\#\# Question: Support for "border-radius" in IE. Does anyone know if/when Internet Explorer will support the \\"border-radius" CSS attribute?\\
    \#\#\# Answer: It is not planned for IE8. See the CSS Compatibility page. Beyond that no plans have been released. \\Rumors exist that IE8 will be the last      version for Windows XP\\
    \#\#\# Feedback: You are obviously wrong, because IE9 is supposed to support CSS3 too, and I dont see IE dying\\ anywhere. Someone pls kill IE. \textbf{[Source: Stack Overflow]}\\
    \midrule
    \#\#\# Question: D.I.Y clay with cornstarch and baking soda.\\
    \#\#\# Answer: Add one cup of cornstarch to 1.5 cups of baking power. Add one cup of water to make a slurry and \\look till it is the right consistency. Cool and use to mold or sculpt. \\
    \#\#\# Feedback: The answer mentions a wrong ingredient. The ingredients must contain baking soda but the \\answer instead mentions baking power. The answer probably meant baking powder but the needed ingredient is \\baking soda.
    \textbf{[Source: Human annotation]} \\
    \bottomrule
    \end{tabular}
    \caption{Examples of our training data collected from Stack Exchange and Human Annotation.}
    \label{tab:finetuning data}
\end{table*}

To fine-tune and evaluate \ourmodel{}, we created a high-quality feedback dataset composing 
of two distinct sets: (1) community feedback, curated from online forums to capture more diverse interactions, and (2) human-annotated feedback, collected on generations across different types of tasks. See examples in Table~\ref{tab:finetuning data}.
Trained on a combination of these datasets, \ourmodel{} shows impressive results, outperforming ChatGPT~\citep{openai2022chatgpt} models on multiple downstream tasks. 
Close inspection of influences of community feedback and human annotated feedback data confirms that the community data is more informative and diverse than human-annotated data, yet leans towards informality. These nuances enable \ourmodel{} to give feedback on diverse tasks, 
while we find that including high-quality human annotated data for finetuning improves model performance. 

We conduct model-based evaluation (using GPT-4~\cite{openai2023gpt4}) as well as human evaluation of the feedback generated by \ourmodel{} comparing against state-of-the-art baselines including Alpaca~\cite{alpaca}, SelFee~\cite{selfee2023}, and ChatGPT~\cite{openai2022chatgpt}. We find that 
critiques from ~\ourmodel{} are generally more preferred over those from other models.
For example, Alpaca tends to give positive feedback to all of model responses, resulting a lot incorrect feedback. SelFee tends to provide vague feedback that cannot pinpoint errors, ignoring model responses or directly answering the question instead of critiquing responses. ChatGPT is more stable across different evaluation settings and we find it does a better job of giving feedback with a correct judgement.

In summary, we develop \ourmodel{}, a new model capable of providing comprehensive critiques of any LLM generated text, which can efficiently enhance their quality. Through rigorous evaluation of generated critiques, we demonstrate the efficacy of \ourmodel{} across a variety of generation tasks. A further key contribution of our work is creating a high-quality feedback dataset, which has the potential to facilitate future research in this area.

\section{Data Collection}
\label{sec:data}

\subsection{Community Critique Data}

We gather feedback data from two community question and answering websites: Stack Exchange\footnote{\url{https://stackexchange.com/}} and the Pushshift Reddit Dataset \citep{baumgartner2020pushshift}.
We consider the title and the sub-title of a post as a question, its top-level comments as answers, and replies to these comments as critiques. 
Everything is associated with a community vote score, calculated by subtracting the total number of downvotes from the total number of upvotes. For the sake of clarity, we will refer to the community vote score as  question score, answer score, and critique score respectively . 

\paragraph{Stack Exchange Data.} 
Stack Exchange contains a network of 173 dedicated Q\&A communities, inclusive of the notable Stack Overflow community, dispensing expert insights in a question and answer format. Our data curation temporarily only focuses on English content. 

\paragraph{Pushshift Reddit Data.} 
Reddit is one of the most popular websites in the world, facilitating the sharing, discussion, and community voting of content within user-generated communities, or ``subreddits''. It hosts nearly 140,000 active subreddits at any given moment. However, there are certain challenges associated with fine-tuning a model on Pushshift Reddit data:
1) a substantial proportion of subreddits function primarily for entertainment rather than serious, informative dialogues. This includes subreddits such as r/funny, r/MadeMeSmile, r/facepalm, and many others;
2) many posts are primarily intended to share information (e.g., news, jokes) instead of posing specific questions, making them unsuitable for fine-tuning.

In light of these limitations, we have selected data from 16 subreddits\footnote{Subreddits used to gather community data: r/AskAcademia, r/AskAnthropology, r/AskBaking, r/askcarguys, r/AskCulinary, r/AskDocs, r/AskEngineers, r/AskHistorians, r/AskHR, r/askphilosophy, r/AskPhysics, r/askscience, r/AskScienceFiction, r/AskSocialScience, r/AskVet, r/explainlikeimfive} from Pushshift Reddit Dataset. 
These communities generally contain more formal discussions, often structured in a Q\&A format. 

\subsubsection{Critique Postprocessing}

We aim to gather instances in the format of a clean question-answer-critique triad. Upon cleaning the community data, we observed that while procuring high-quality Q\&A pairs was relatively straightforward, a significant portion of the data did not include suitable critiques. We consider a critique to be valid in two scenarios.

- \textbf{Case \#1:} The answer is largely accurate, and the critique offers recommendations for further refinement or enhancement.

- \textbf{Case \#2:} The answer contains inaccuracies, which the critique explicitly highlights.

We employ a number of techniques to curate valid critiques from community data. First, we filter out invalid critique data, such as joke sharing and follow-up questions that fail to provide feedback. To do this, we use two methods. \textbf{(1) Keyword filtering:} We keep examples that contain certain keywords matching the above two cases\footnote{For Case \#1, keywords we use for filtering are: ``not wrong'', ``agree'', ``absolutely'', ``indeed'', ``agreed'', ``exactly what I think'', ``that's right'', ``not what I think'', ``you're right'', ``you are right'', ``that is right''; For Case \#2, keywords we use are: ``wrong'', ``incorrect'', ``not agree'', ``not right'', ``disagree'', ``can't agree'', ``beg to differ'', ``that's not my view''.}. 
\textbf{(2) User edit history:} In additional to keyword filtering, we collect critiques if users edit their answer after posting the critique. This helps us identify examples where the critique leads to a modification of the original answer.

Next, we note that even critiques that follow the appropriate format may be inaccurate. For instance, a critique might wrongly challenge a true fact from an answer. To further refine our data and ensure high quality of critiques, we incorporate additional filters linked with community vote scores. 
In Case \#1, where the answer is primarily correct, we omit data where the answer score is lower than 10 and the critique score is lower than 2. This filter ensures we only select instances where the initial answer is largely approved by the community, and the critique has received some level of endorsement.
In Case \#2, where the answer includes inaccuracies, we focus on data where the critique score surpasses the answer score, and the critique score itself is higher than 2. This selection process ensures we are considering instances where the critique, indicating an error in the initial answer, has garnered more community approval than the answer itself.

Lastly, we impose several additional filters to further refine our dataset:
1) To maintain diversity, we retain only one instance per post, choosing the one with the highest critique score;
2) To manage offensive language in community data, we incorporate a profanity check\footnote{\url{https://pypi.org/project/alt-profanity-check/}} and eliminate any comments or feedback with a profanity score lower than 0.8;
3) Given that our model is a text-only model, we filter out instances that contain URLs, images, or videos;
4) We identify and remove comments that pose further questions to the original question rather than the original answer to preserve the integrity of the Q\&A format.

\subsection{Human Data Collection}

\paragraph{Public dataset selection.}
To collect natural language feedback from a diverse set of contexts and outputs, we select $8$ popular language-understanding and entailment datasets that require complex reasoning and have step-by-step explanations to arrive at the final answer, as well as two summarization datasets. 
Specifically, we choose 
Entailment Bank (deductive reasoning) \citep{entailmentbank2021}, 
Proofwriter (logical reasoning) \citep{tafjord2020proofwriter}, 
GSM8K (arithmetic reasoning) \citep{cobbe2021gsm8k}, 
PIQA (physical reasoning) \citep{bisk2020piqa}, 
CosmosQA (commonsense reasoning) \citep{huang2019cosmos}, 
ECQA (commonsense reasoning) \citep{aggarwaletal2021ecqa}, 
e-SNLI (deductive and commonsense reasoning) \citep{camburu2018snli}, 
and Adversarial NLI (adversarial entailment) \citep{nie2019adversarial}\footnote{For entailment datasets such as e-SNLI and Adversarial NLI, only prompts with a target label of ``entailment'' or ``contradiction'' are used, i.e., prompts labelled with ``neutral'' are removed. This is because it is difficult for our critique model to learn from neutral examples to give useful feedback.}. 
For summarization tasks, we choose two datasets that are of relatively high quality: GPT-3 summarization \citep{goyal2022news} and DeFacto \citep{liu2022improving}. 
Only data from the training sets is used for human annotation  
\footnote{To diversify the human data, we also add one or two examples from the training set of each of these datasets: SIQA \citep{sap2019socialiqa}, BoolQ \citep{clark2019boolq}, the challenge set and the easy set from ARC \citep{clark2018think}, the middle school set and the high school set from RACE \citep{lai2017large}, HellaSwag \citep{zellers2019hellaswag}, Natural Question \citep{kwiatkowski2019natural}, WinoGrande \citep{sakaguchi2021winogrande}, TriviaQA \citep{joshi2017triviaqa}.}.

\paragraph{Data construction for human annotation.}
In order to collect high quality natural language feedback, for each question we provide a context, a correct output, and a candidate output, and ask annotators to give feedback on whether there are any errors in the candidate output. The correct output is provided to help annotators identify errors more easily. 
Except for GPT-3 summarization, in which we choose the summary best preferred by human raters, all datasets contain a gold answer to the question, i.e., the correct output. 
PIQA, CosmosQA, and DeFacto have candidate outputs in the original datasets. 
For GPT-3 summarization we choose the summary least preferred by human raters as the candidate output. 
For other datasets, to increase the possibility of obtaining candidate outputs that have reasonable errors, we prompt LLaMA-65B \citep{touvron2023llama} or LIMA-30B \citep{zhou2023lima} with zero-shot or few-shot in-context examples to obtain step-by-step reasoning sequences. 
We do a manual inspection to ensure the quality of the generated candidate outputs. 
Examples from each dataset and the templates used to prompt LLaMA and LIMA in order to obtain the candidate outputs are shown in Table~\ref{tab:human_data_dist}.

\paragraph{Annotation specifics.}
We selected an independent vendor (RWS Moravia), utilizing expert reviewers instead of crowd-sourcing, due to the cognitively demanding nature of our annotation task which necessitates meticulous and nuanced annotations. 
Each example is annotated by one expert with human-in-the-loop quality assessment.\footnote{Each example is quoted for \$8 by RWS Moravia. Annotators are native English speakers with strong writing skills and with a high school degree and above. 
}
We define different error types in our taxonomy in Table~\ref{tab:error_types}. 
More details about the instruction and guidelines of human annotation can be found in Appendix~\ref{appendix:human}. 

\paragraph{Postprocessing.}
To ensure high quality data, we conduct the following postprocessing.
First, we remove examples flagged with ``Errors in the correct output'' and ``The context is too complex to work on''.  
We also remove feedback on error types of ``Redundancy'' and ``Consistency with context'' because we find that the feedback on these two error types are not helpful. 
As our ultimate goal is to collect feedback in natural language, we concatenate the feedback from different error types of an example into a paragraph using natural words such as ``Firstly,'' ``Secondly,'' and ``Besides'' if there exists feedback for more than one error type in a single example. 
We end up with $1$,$317$ high quality examples in total. 
Table~\ref{tab:number_errortypes} presents the distribution of collected examples and the identified error types from each dataset used in human annotation. 

\section{The \ourmodel{} Model}
\label{sec:training}
We train~\ourmodel~ with LLaMA-7B ~\cite{touvron2023llama} as the base model. We use AdamW~\cite{DBLP:journals/corr/abs-1711-05101} as our optimizer with $\beta_1 = 0.9$, $\beta_2 = 0.95$ and weight decay of $0.1$. We use a learning rate 1e-5 with 2000 warmup steps, and set the batch size to 64 and maximum sequence length to 2048. The training data is formatted using the same template where we use ``\#\#\# \{field name\}'' to separate different fields as shown in Table~\ref{tab:finetuning data}.

We keep checkpoints for each $50$ steps with a total of $3,000$ steps. We manually inspect whether the generated feedback can identify errors or come up with constructive suggestions on a held-out set of $20$-examples and select $3$ best checkpoints. 
After that, we use the GPT-4 evaluation protocol (as described in ~\ref{sec: GPT4 eval}) to pick the best checkpoint on the held-out set.

\section{Evaluating Feedback}
\label{sec:eval}
\begin{figure}[t]
\fbox{\begin{minipage}{19.5em}
\scriptsize
Your task is to evaluate the feedback on a model-generated answer. You'll receive a question, an answer, and feedback about this answer. Give a score based on the feedback's quality.\\\\
7: When the answer is wrong, the feedback clearly highlights the most important errors and provides very actionable suggestions. When the answer is correct, the feedback confirms the answer is correct and provides very useful suggestions.\\
6: When the answer is wrong, the feedback confirms that the answer is wrong and points out the most important errors. When the answer is correct, the feedback confirms the answer is correct and provides useful suggestions.\\
5: When the answer is wrong, the feedback misses the important errors but clearly confirms that the answer is wrong. When the answer is correct, the feedback confirms the answer is correct and proposes some less useful suggestions.\\
4: The feedback has a correct judgement of the answer (e.g., states correct answer is correct or states wrong answer is wrong).\\
3: The feedback is vague about whether or not the answer is correct. Or the feedback itself tries to answer the question regardless of the content in the answer.\\
2: The feedback has a wrong judgement of the answer (e.g., states correct answer is wrong or states wrong answer is correct).\\
1: The feedback is completely random text or simply repeats the answer.\\\\
First, please check whether the feedback has correct or incorrect judegement (correct judgement means the answer is correct, feedback confirms the correctness. Or if the answer is incorrect, the feedback confirms the incorrectness.)\\\\ Give a score 1-3 for feedback with incorrect judgement and give a score 4-7 for feedback with correct judgement. \\\\
Start your answer with the score.
\end{minipage}}
\caption{Instruction for GPT-4 and human evaluation to rate each feedback on a 1-7 likert scale.}
\label{fig:likert_instruction}
\end{figure}

To test the ability of \ourmodel{} to critique model generations, we compare it against a series of state-of-the-art language models including Alpaca-7B~\cite{alpaca}, SelFee-7B~\cite{selfee2023} and ChatGPT (GPT-3.5 Turbo). We conduct both human evaluation and automatic evaluation by using GPT-4~\cite{openai2023gpt4} as an evaluator. Through extensive experiments across multiple downstream tasks, we show our model can effectively critique model generations and propose suggestions, outperforming Alpaca and SelFee while achieving performance that is competitive with ChatGPT.

\subsection{Evaluation Data}
To get a broad coverage of NLP domains, we carefully select 6 public datasets for evaluation: AlpacaFarm~\cite{dubois2023alpacafarm}, FairEval~\cite{wang2023large}, CommonsenseQA~\cite{talmor-etal-2019-commonsenseqa}, OBQA~\cite{Mihaylov2018CanAS}, PIQA~\cite{Bisk2020}, TruthfulQA~\cite{lin2021truthfulqa}. These 6 datasets cover a wide range of topics and reasoning skill-sets, including commonsense reasoning, physical reasoning, mathematical reasoning, etc. We sample 50 instances from the validation or test set of each dataset, resulting in a total of 300 instances in the final evaluation set. We further sample 20 instances from each dataset to be used for our ablation studies. 

\textbf{Data contamination.} The issue of data contamination has been widely discussed along with ChatGPT and related models. To prevent data contamination in our evaluation, we further develop a new test set CritiqueEval, containing 52 Pushshift questions posted from June 2022 to June 2023\footnote{The current version of ChatGPT contains training data up to Sep. 2021. Our training set contains training data up to Dec. 2021.}. We manually select questions from Pushshift Reddit dataset with highest voting score, and for each question, we collect an answer with the highest vote score and another answer with the lowest vote score\footnote{community vote score from Pushshift Reddit dataset is calculated by the total number of upvotes minus the total number of downvotes.}. 
In summary, our evaluation data in total consists of 352 instances from public datasets as well as from Pushshift Reddit dataset. 

\begin{figure}[t]
\fbox{\begin{minipage}{19em}
\scriptsize
You are a helpful and precise assistant for checking the quality of the feedback.\\
Two pieces of feedback have been provided for the same response to a particular question. Which one can point out key errors contained in the answer or provide constructive suggestions?\\
\\
Feedback 1: ...\\
Feedback 2: ...\\
\\
Please choose from the following options.\\
A: Feedback 1 is significantly better.\\
B: Feedback 2 is significantly better.\\
C: Neither is significantly better.
\end{minipage}}
\caption{Instruction for GPT-4 and human evaluation to pick a better critique out of two.}
\label{fig:pairwise_instruction}
\end{figure}

\subsection{Baseline Models}
We compare \ourmodel{} against the following state-of-the-art baselines:
\begin{itemize}[leftmargin=*]
\itemsep0em 
    \item \textbf{ChatGPT}, one of the most powerful chatbots developed by OpenAI.
    \item \textbf{Alpaca-7B}, a LLaMA-7B model finetuned on 52K instruction-following data obtained from ChatGPT. We include this model since details about the language model behind ChatGPT are not revealed.
    \item \textbf{SelFee}, a finetuned LLaMA-7B model but curated for self-feedback and self-revision generation. The training set for this model includes 178K training instances from five different data sources, obtained by querying ChatGPT to get original generations, feedback and revisions. Note that using exact matching on questions, we observe some overlap between our evaluation data and SelFee's training data, namely 33 out of 50 instances from AlpacaFarm, 6 out of 50 instances from TruthfulQA, 1 out of 50 instances from OBQA and 1 out of 50 instances from PIQA.  
\end{itemize}

\subsection{GPT4 Evaluation}
\label{sec: GPT4 eval}
Recent work has used GPT-4~\cite{openai2023gpt4} 
to assess the quality of model-generated text~\cite{liu2023geval, wang2023far, zhou2023lima}. We follow the same approach to prompt GPT-4\footnote{\texttt{GPT-4-0613} is used in our experiments.} to measure to what extent our model can critique answers generated by LLMs.

\subsubsection{Absolute Likert Score}
We ask GPT-4 to grade each feedback on a 1--7 likert scale mainly based on whether the feedback can point out errors in the answer or confirm the answer is correct when there are no errors. We provide the instruction shown in Figure~\ref{fig:likert_instruction} as the system message to GPT-4. Note GPT-4 can make mistakes when improper instruction is provided. We select the instruction shown in Figure~\ref{fig:likert_instruction} through manual inspection, conducted by testing various instructions on a series of feedback examples and assessing if the grading by GPT-4 aligns with human judgement. We provide a few examples in Appendix~\ref{sec: instruction_derivation} to show how GPT-4 performs under this chosen instruction. Despite this careful instruction-selection process, GPT-4 may still be misaligned with human judgment, as evinced in Table~\ref{tab:gpt4_failure}.

\subsubsection{Pairwise Comparison}
We also ask GPT-4 to pick the better feedback given two candidate feedback. To do this, we provide the instruction shown in Figure~\ref{fig:pairwise_instruction}. In this setting, instead of giving detailed instructions, we directly ask GPT-4 to pick the better feedback according to whether it can identify errors when the response is wrong or provide useful suggestions when the response is correct but needs improvement.

{\small
\begin{table}[]
\centering
\scalebox{0.7}{
\begin{tabular}{|p{1.5cm}|p{8cm}|}\hline
Question               & Why the elderly body odor is so strong? \\ \hline
Response               & As our skin matures, its natural antioxidant protection declines. This decline results in greater oxidation of lipid acid. When lipid acid is oxidized, the chemical compound nonenal is produced, giving off the “old people smell” that many of us are familiar with.                      \\ \hline
Feedback               &        The response gives a logical explanation to the question, using simple and understandable language. However, it could benefit from proper punctuation and grammar. For example, it would be more accurate to say that the chemical compound is nonanal, rather than nonenal, and the sentence could be rewritten as follows: 

As our skin matures, its natural antioxidant protection declines. This decline results in greater oxidation of lipid acid. When lipid acid is oxidized, the chemical compound nonanal is produced, giving off the “old people smell” that many of us are familiar with.                  \\ \hline
Related Knoledge from Wikipedia & One study suggested that old person smell may be the result of 2-nonenal, an unsaturated aldehyde which is associated with human body odor alterations during aging; however, there are other hypotheses. Another study failed to detect 2-nonenal at all, but found significantly increased concentrations of benzothiazole, dimethylsulphone, and nonanal on older subjects. \\ \hline
\end{tabular}}
\caption{Challenges in human evaluation. There are cases where deep domain knowledge is required to evaluate various feedback.}
\label{tab:human_eval_difficulty}
\end{table}
}

\subsection{Human Evaluation}
As language models may be biased evaluators 
\cite{aiyappa2023can, wang2023far}, we further conduct human evaluation. Similarly to the GPT-4 evaluation, we ask each crowd worker to rate feedback on a 1--7 likert score. For a fairer comparison, we present the question, answer and feedback generated by different models together to annotators. Note that GPT-4 rates each feedback independently, whereas crowd workers are required to compare the quality of multiple feedback and assign scores accordingly. In this way, we encourage crowd workers to rank multiple feedback for the same question-answer pair.

\begin{figure}[t]
    \centering
    \includegraphics[width=0.9\linewidth]{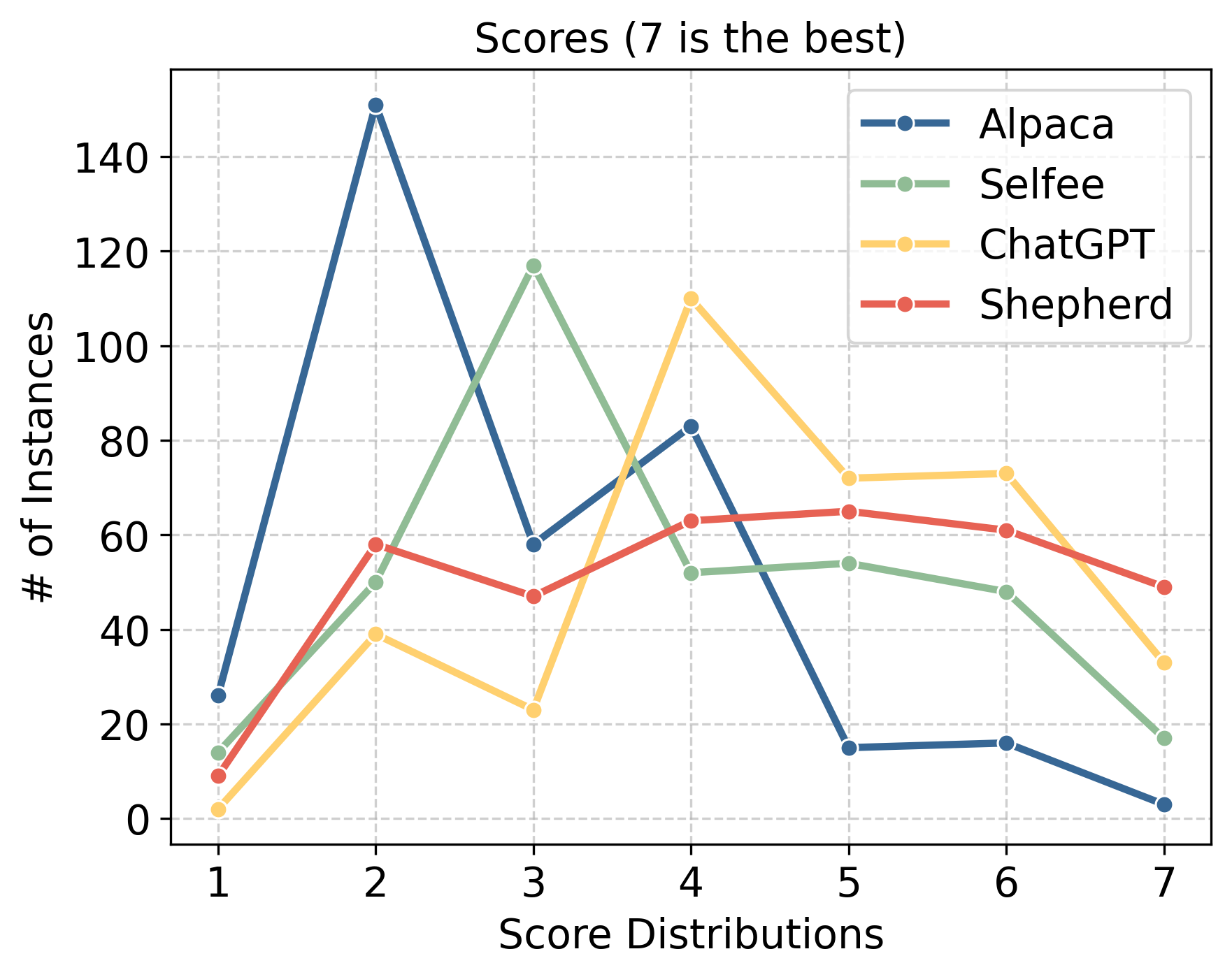}
    \caption{Score distribution from human evaluation. Scores ranging from 1--2 means the model fails to give correct judgements. A score of 7 means the feedback not only gives a correct judgement of the answer but also provides very helpful feedback for improving the answer.}
    \label{fig:human_eval_curves}
\end{figure}

Questions in our evaluation data may require specific domain knowledge. Even though we invite expert annotators to do human evaluation who carefully search online whenever they need help, human evaluation is still a challenging task. Table~\ref{tab:human_eval_difficulty} shows one example that requires deep medical or biology knowledge.

\section{Results}
\label{sec:results}
\subsection{Pairwise Comparison}
\todo{There are 4 experiment results: Human pairwise comparison, GPT4 pairwise comparison, Human likert score, GPT4 likert score. It seems this paragraph is about GPT4 pairwise comparison?}

We first analyze whether \ourmodel{} can generate better feedback compared to other competing models. We present pairwise comparison results in Figure~\ref{fig:pairwise human} and Figure~\ref{fig:pairwise gpt4}, using GPT-4 and human evaluation, respectively. In both evaluation settings, \ourmodel{} significantly outperforms Alpaca and consistently outperforms SelFee. Note that both \ourmodel{} and SelFee are finetuned LLaMA-7B models, however, SelFee was finetuned on a dataset of 178K examples (some of which overlapped with our evaluation data) whereas \ourmodel{} was only finetuned on a dataset of 8K. We also see \ourmodel{} has slightly better performance than ChatGPT according to GPT-4 evaluation, and on par performance in human evaluation. 

Detailed comparison results for each task are presented in Table~\ref{tab:win rate by GPT-4} and Table~\ref{tab:win rate by human}. The performance varies across tasks but overall aligns across the two evaluation settings. It is important to note that \ourmodel{}  consistently generates better feedback on CritiqueEval, a dataset which has never been seen by ChatGPT or LLaMA models. This demonstrates both the critique and generalization abilities of our \ourmodel{} model. Some qualitative examples are included in Table~\ref{tab:qualitative}.

\begin{table*}[ht]
    \centering
    \small
    \begin{tabular}{lcccccccc}
    \toprule
      {\diagbox{\emph{Win Rate}}{\emph{Test Data}}} & AlpacaFarm & FairEval & CosmosQA & OBQA & PIQA & TruthfulQA & CritiqueEval & Avg.\\
      \midrule
      \ourmodel{}  vs. Alpaca & 78.0 & 90.0 & 82.0 & 87.0 & 91.0 & 90.0 & 92.3 & 87.0 \\
      \midrule
      \ourmodel{}  vs. SelFee & 52.0 & 55.0 & 58.0 & 51.0 & 46.0 & 57.0 & 51.9 & 53.0\\
      \midrule
      \ourmodel{}  vs. ChatGPT & 55.0 & 65.0 & 47.0 & 47.0 & 66.0 & 48.0 & 61.5 & 56.0\\
      \bottomrule
    \end{tabular}
    \caption{Win rate (\%) by GPT-4 evaluation. Overall, ~\ourmodel{} outperforms all competing models including ChatGPT. On the other hand, models perform differently across tasks, e.g.~\ourmodel{}  is better than ChatGPT on PIQA task but not TruthfulQA. }
    \label{tab:win rate by GPT-4}
\end{table*}

\begin{table*}[ht]
    \centering
    \small
    \begin{tabular}{lcccccccc}
    \toprule
      {\diagbox{\emph{Win Rate}}{\emph{Test Data}}} & AlpacaFarm & FairEval & CosmosQA & OBQA & PIQA & TruthfulQA & CritiqueEval & Avg.\\
      \midrule
      \ourmodel{}  vs. Alpaca& 67.0 & 76.0 & 76.0 & 69.0 & 69.0 & 82.0& 86.5 &72.4 \\
      \midrule
      \ourmodel{}  vs. SelFee & 55.0 & 59.0 & 62.0 & 52.0& 53.0 & 68.0& 68.3 & 59.7\\
      \midrule
      \ourmodel{}  vs. ChatGPT & 38.0 & 59.0 & 39.0 & 49.0& 43.0 & 52.0& 66.3 & 49.6\\
      \bottomrule
    \end{tabular}
    \caption{Win rate (\%) by human evaluation. \ourmodel{} significantly outperforms Alapca and SelFee, and on average it ties with ChatGPT.}
    \label{tab:win rate by human}
\end{table*}

\begin{table*}[ht]
    \centering
    \small
    \begin{tabular}{lccccccccc}
    \toprule
      {\diagbox{\emph{Model}}{\emph{Test Data}}} & AlpacaFarm & FairEval & CosmosQA & OBQA & PIQA & TruthfulQA & CritiqueEval & Avg.\\
      \midrule
      Alpaca 7B & 2.96 & 2.90 & 2.61 &  2.91&  2.88& 2.75 & 2.73 &  2.91 \\
      \midrule
      SelFee 7B & 3.54 & 3.90 & 3.33 & 3.69 &  3.82& 3.57 & 3.38 &  3.84\\
      \midrule
      ChatGPT & 4.56 & 4.31 & 4.28 &  4.30& 4.43 & 4.27 & 4.04 & 4.59\\
      \midrule
      \ourmodel{} 7B & 4.38 & 4.75 & 4.36 & 4.58& 4.44 & 4.66 & 4.65 & 4.41 \\
      \bottomrule
    \end{tabular}
    \caption{Likert score on 1-7 scale given by Human. Alpaca and SelFee perform significantly worse than ChatGPT and Shepherd. Overall ChatGPT and Shepherd have similar performance while ChatGPT is generally better on exsiting public datasets and Shepherd is significanly better on CritiqueEval.}
    \label{tab:likert score by Human}
\end{table*}
\begin{table*}[ht]
    \centering
    \small
    \begin{tabular}{lccccccccc}
    \toprule
      {\diagbox{\emph{Model}}{\emph{Test Data}}} & AlpacaFarm & FairEval & CosmosQA & OBQA & PIQA & TruthfulQA & CritiqueEval & Avg.\\
      \midrule
      Alpaca 7B & 4.83 & 4.62 & 4.28 & 4.84 & 4.72 & 4.24 & 5.27 & 4.69  \\
      \midrule
      SelFee 7B & 5.28 & 5.60 & 5.12 & 5.54 & 5.60 & 5.71 & 5.60 & 5.49 \\
      \midrule
      ChatGPT & 6.54 & 6.46 & 5.98 & 6.10 & 6.51 & 6.16 & 6.00 & 6.25\\
      \midrule
      \ourmodel{} 7B & 5.38 & 5.88 & 4.98 & 5.08& 5.38 & 5.42 & 5.44 & 5.37 \\
      \bottomrule
    \end{tabular}
    \caption{Likert score on 1-7 scale given by GPT-4. Alpaca performs worse than other models but gets a high score 4.69. This does not align with the aforementioned evaluation results where Alpaca generates mostly wrong feedback. On the other hand, the gap between different models are also relatively smaller.
    We think it is partially due to the issue that GPT-4 is not capable of strictly following instructions to do evaluation in this setting.}
    \label{tab:likert score by GPT4}
\end{table*}

\subsection{Score Distribution}
We also look into the distribution of scores using human evaluation. As shown in Figure~\ref{fig:human_eval_curves}, most feedbacks from Alpaca have an incorrect judgement of the provided answer (score 2). SelFee tends to generate vague feedback or directly answers the question instead of critiquing (score 3). Most feedbacks from ChatGPT have a correct judgement (score 4). Compared to ChatGPT, ~\ourmodel{} provides more helpful suggestions to improve the answer (more feedback with score 7) although it tends to generate feedback with wrong judgements (more feedback with score 2). 

\subsection{Likert Score}
We include results using likert scores annotated by GPT-4 in Table~\ref{tab:likert score by GPT4} and likert scores annotated by human annotator in Table~\ref{tab:likert score by Human}. Compared to pairwise comparison by GPT-4, we find the likert score evaluation by GPT-4 is less convincing with the following reasons:

(i) When GPT-4 is asked to rate each feedback independently, it tends to give a high rating for all feedback. For example,  Alpaca obtains an average score of 4.7, meaning most feedback have a correct judgment of the response, while human gives an average score of 2.91, meaning a great proportion of feedback have the incorrect judgement. Evaluation results obtained by using GPT-4 conflict with the observations from the human evaluation.

(ii) The likert score assigned by GPT-4 appears to be favoring a specific format. According to human evaluation, SelFee frequently provides minor and general suggestions, such as adding more examples to better illustrate the answer, even when an ample amount of examples has already been included in the response. In such cases, GPT-4 routinely give higher scores. However, human annotators are typically not misled by this pattern. This results in a discrepancy where GPT-4 assigns high scores to the Selfee model, while human evaluators give significantly lower scores.

(iii) We also see that the performance gap between these models is relatively small, making it difficult to distinguish good and bad models. 

Overall, we recommend using GPT-4 to rate in a pairwise comparison style, which aligns well with human evaluation. 
\paragraph{Limitation of GPT-4 evaluation.} We find that GPT-4 evaluation can be unstable and inaccurate in two cases.
\textit{(i)} One instance of GPT-4's evaluation failure is illustrated in Table~\ref{tab:gpt4_failure}. Even when the correct answer is provided, GPT-4 fails to recognize whether feedback has a correct judgement.
\textit{(ii)} We observed that GPT-4 exhibits some degree of knowledge barrier, affecting its ability to score accurately. This is demonstrated in Table~\ref{tab:gpt4_failure_2}, where \ourmodel{} identifies the error in the answer, whereas all other models perceive the answer as correct. Owing to the inherent knowledge barrier, GPT-4 assigns a low score to \ourmodel{} while rating the other models highly, although only \ourmodel{} identifies the error in the answer.

\subsection{Data Matters}
\begin{figure}[t]
    \centering
    \includegraphics[width=\linewidth]{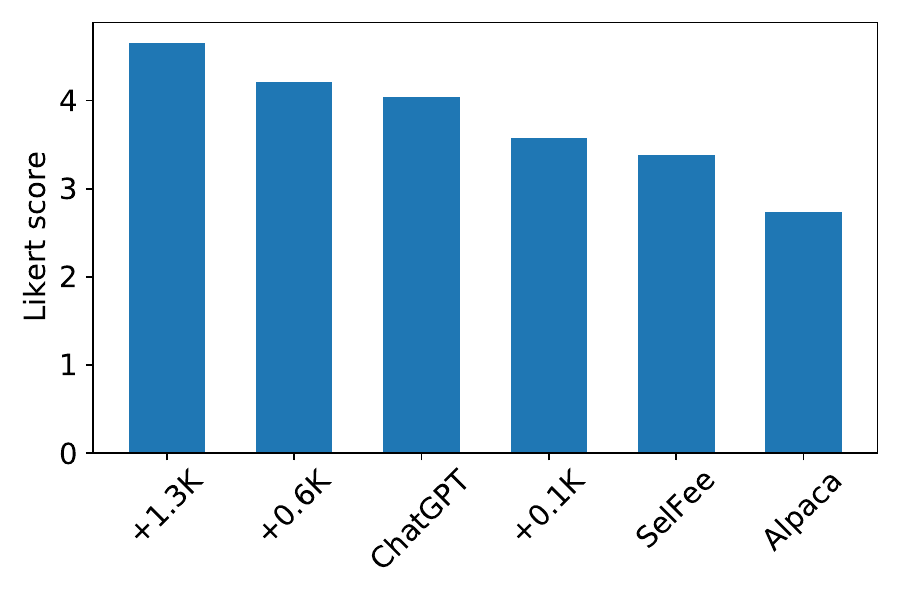}
    \caption{We analyze different critique models' performance on CritiqueEval set by GPT-4 evaluation (averaged across multiple instructions). +0.1K/+0.6K/+1.3K means a \ourmodel{} model finetuned on all community data and 0.1K/0.6K/1.3K human data. With more human annotated data included, we are able to build better critique models.
    }
    \label{fig:increase data}
\end{figure}

In this experiment, we explore how much data is sufficient to build a good critique model. We show results in Figure~\ref{fig:increase data}. We find that when data is of high quality, more finetuning data leads to a better critique model. Compared to SelFee which uses much more finetuning data, we demonstrate our high quality community data and human annotated data are more useful for building critique models.

\begin{table*}[ht]
    \centering
    \scriptsize
    \begin{tabular}{l|l}
    \toprule
    \makecell[l]{Question and \\Response} & \makecell[l]{Question: What constitutes a frog's diet?\\
    Response: A frog's diet consists of insects and other small 
    invertebrates. Fruits, vegetables, and nectar are also eaten by some \\species of frogs.\\
    Ground truth from OBQA: it chomps on insects}\\
    \midrule
    \makecell[l]{Feedback from \\ Alpaca} & \makecell[l]{The answer is correct and provides a comprehensive overview of what constitutes a frog's diet. It also includes important information \\about some species of frogs that also eat fruits, vegetables, and nectar. Overall, a great job!}\\
    \midrule
    \makecell[l]{Feedback from \\ SelFee} & \makecell[l]{The answer is accurate, but it could benefit from more specific details. For example, it could mention that certain species of frogs are \\more carnivorous than others and may even eat small vertebrates like fish or birds. Additionally, it could mention that some frogs \\have  been known to eat plant material in addition to insects and invertebrates, but this is not a significant part of their diet. Overall, the \\answer is a solid 8/10, but could be improved with more specific details. Revision is needed.}\\
    \midrule
    \makecell[l]{Feedback from \\ ChatGPT} & \makecell[l]{The response to the question is accurate and provides a good overview of what constitutes a frog's diet. It is beneficial to include the \\additional information about some species of frogs also eating fruits, vegetables, and nectar. The answer is clear and concise, making it\\ easily understandable for anyone who may be unfamiliar with the topic. Overall, well done!}\\
    \midrule
    \makecell[l]{Feedback from \\ \ourmodel} & \makecell[l]{The response provided is not accurate and should be revised. A frog's diet varies depending on the species of frog. For example, some \\species of frogs are carnivorous and prey on small insects and arthropods, such as worms and spiders. Other species of frogs are \\herbivorous and primarily eat plants, fruits, and vegetables. Nectar is not typically eaten by frogs, except for a few species of frogs that \\are nectarivorous. A more accurate response would be: A frog's diet varies depending on the species of frog. For example, some species \\of frogs are carnivorous and prey on small insects and arthropods, such as worms and spiders. Other species of frogs are herbivorous and\\ primarily eat plants, fruits, and vegetables. Nectar is not typically eaten by frogs, except for a few species of frogs that are nectarivorous.}\\
    \midrule\midrule
    \makecell[l]{Question and \\Response} & \makecell[l]{Question: What are some great financial investments with no risk at all? \\Response: Investing in the stock market,     mutual funds, bonds, and real estate are all great financial investments with no risk at all.}\\
    \midrule
    \makecell[l]{Feedback from \\ Alpaca} & \makecell[l]{This response is well        structured and provides a comprehensive overview of the various types of investments that have the potential to be \\profitable with minimal risk. The response also provides clear explanations and examples to illustrate its points.}\\
    \midrule
    \makecell[l]{Feedback from \\ SelFee} & \makecell[l]{The answer is not entirely     accurate. While it is true that investing in the stock market, mutual funds, bonds, and real estate can provide good \\returns, there is always a level of risk involved. It is important to do thorough research and understand the potential risks and benefits of each \\investment before making a decision. Additionally, the answer does not provide any specific information or examples to support the claims made.}\\
    \midrule
    \makecell[l]{Feedback from \\ ChatGPT} & \makecell[l]{This response is               incorrect. There is no such thing as a financial investment with no risk at all. Even low-risk investments such as bonds or\\ mutual funds have some level of risk associated with them. It is important to thoroughly research and understand the level of risk involved \\with any investment before making a decision.}\\
    \midrule
    \makecell[l]{Feedback from \\ \ourmodel} & \makecell[l]{The response is not            accurate and does not address the question fully. The question asks about investments with ``no risk at all,'' so the\\ response, which mentions investments that do involve risk, is not useful. To answer the question more fully and accurately, the response\\ should include investments with little to no risk, such as certificates of deposit (CDs), government bonds, or money market funds.}\\
    \bottomrule
    \end{tabular}
    \caption{Qualitative examples of feedback generated by different models. }
    \label{tab:qualitative}
\end{table*}

\section{Related Work}
\label{sec:related_work}
\textbf{Evaluation of LLMs.}
Evaluation of natural language generations is an active area of research. Human evaluation of model generations which is a reliable form of evaluation is an expensive, domain specific, and time-consuming process. Existing automatic metrics including word overlap or embedding-based similarity with reference outputs are also not equipped to measure logical inconsistencies, factuality or alignment with human judgement \cite{golovneva2022roscoe, wang2023large}. To address these limitations, recent work investigated the use of LLMs for reference-free evaluation of language model generations \cite{chiang2023vicuna, peng2023instruction, sun2023principledriven, zhou2023lima, wang2023PandaLM, liu2023geval}. Other work~\cite{wang2023far, wang2023large} however points out that the biases in LLMs can lead to unfair evaluations.

\vspace{5pt}
\noindent
\textbf{Critique models.} Recent work on improving LLMs performance has explored an iterative natural language feedback and refinement approach. Self-Improve~\cite{selfimprove} uses high-confidence rationale-augmented answers generated by a 540B-parameter LLM for fine-tuning it and shows improvements on general reasoning abilities of the model. \citeauthor{saunders2022selfcritiquing} (\citeyear{saunders2022selfcritiquing}) finetune a LLM to write natural language critiques for both model generated and human written text and argue that larger models can write more helpful critiques and can use these critiques to self-refine. Self-Refine~\cite{selfrefine} uses a single LLM for generation, providing feedback, and refinement. 
Critic~\cite{critic} proposes interacting with tools for evaluating certain aspects of text, and using the feedback for refinement. SelFee~\cite{selfee2023} collects generations, feedback and revised generations from ChatGPT and finetunes LLaMA models to build a critique model. Self-Correct~\cite{self-correct} decouples the generator from the corrector and shows generations can be improved even when the corrector is much smaller than the base generator.

\section{Conclusion}
\label{sec:conclusion}
We introduce a new model to critique large language model generations. Through extensive experiments across multiple datasets and different evaluation setting, we demonstrate our model can effectively critique answers, reaching performance on par with ChatGPT. With LLMs adopted in more and more real life applications, we believe it is important to develop automatic mechanisms to inspect model generations. Our critique model~\ourmodel~can be very helpful to improve generation quality and reduce hallucinations.

\section*{Limitations}
In this paper, we evaluate models' critique ability through both automatic evaluation and human evaluation. 
Despite our efforts to experiment with a substantial number of data examples and employ human annotators to the best of our financial capabilities, it is evident that the paper could benefit from further enhancement. This can be achieved through the execution of a more meticulous analysis carried out with increased number of annotators and a broader array of diverse tasks.

\section*{Acknowledgements}
We thank Justine Kao, Carleigh Wood, and RWS Moravia vendor for suggestions on human data collection. 

\bibliography{anthology,custom}
\bibliographystyle{acl_natbib}
\newpage
\appendix
\label{sec:appendix}
\section{Human Data Collection Guidelines}
\label{appendix:human}

\begin{figure*}
    \centering
    \includegraphics[width=0.8\textwidth]{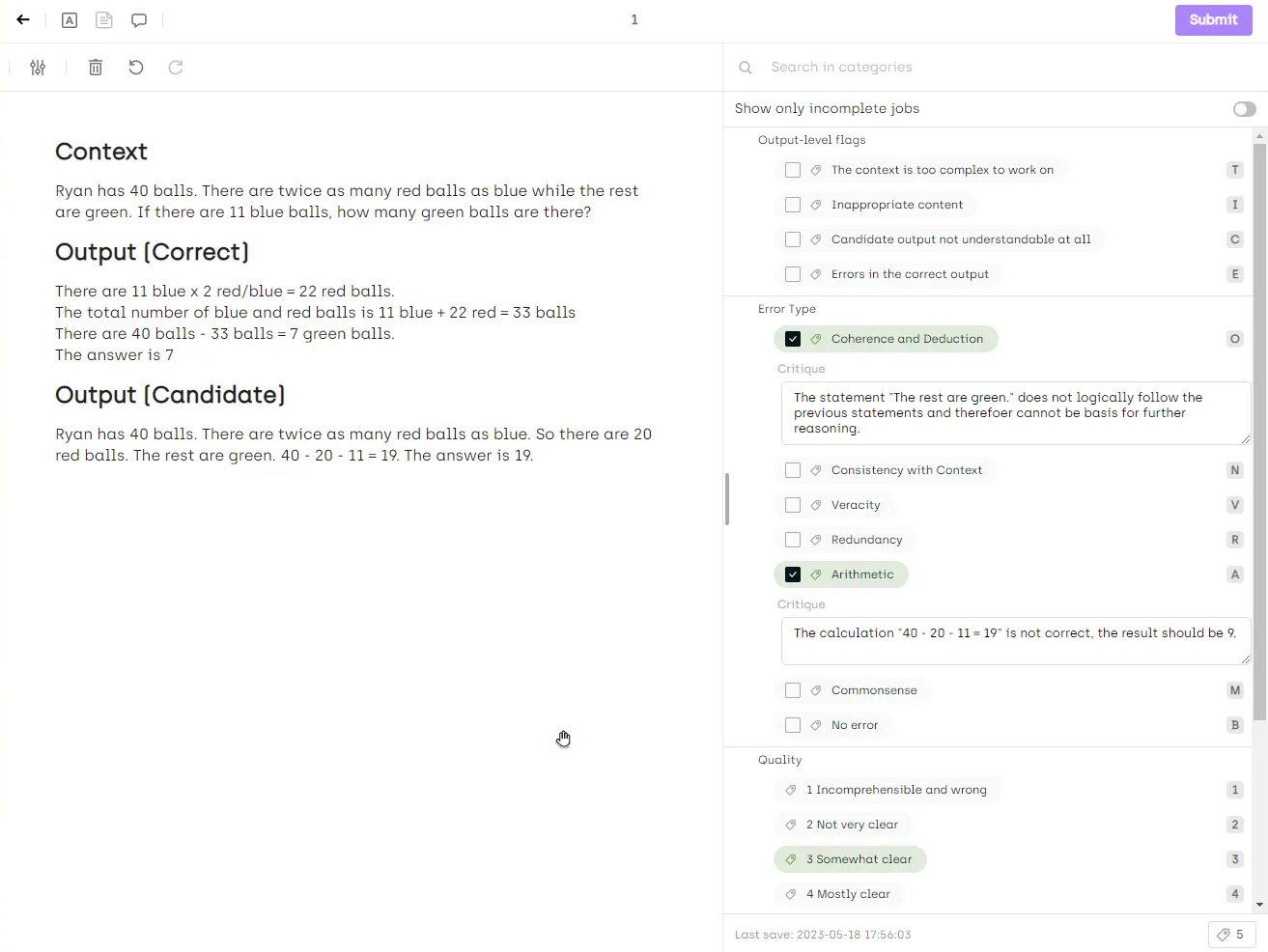}
    \caption{Data collection interface for human annotated data. 
    }
    \label{fig:vendor_ui}
\end{figure*}

In order to collect high quality feedback in natural languages, for each question we provide a context, a correct output, and a candidate output, and ask annotators to give feedback on whether there is any error in the candidate output. The correct output is provided to help annotators identify errors in the candidate output more easily. 
The overview and high level instruction for annotation is provided in Figure~\ref{fig:guide}. 
We define different error types in our taxonomy in Table~\ref{tab:error_types} and examples of questions and feedback for each error type in Table~\ref{tab:human_examples}. 
To ensure the quality of the collected data, four output-level flags are introduced to identify examples are too complicated, contain inappropriate content, contain candidate outputs that are incoherent, and contain ground truth outputs that have errors. 
The data collection interface is provided in Figure~\ref{fig:vendor_ui}.

\begin{table*}[p]
\begin{center}
\renewcommand{\arraystretch}{1.2}
\scalebox{0.75}
	    {
\begin{tabular}{@{}p{100pt}p{500pt}@{}} 
\toprule
{Dataset}
& {Example prompt in template} \ \  
\\ \midrule

Entailment Bank  
&   
\textcolor{red}{Here is a question:}\par
Most birds sit on their eggs until they hatch. Which of these is the most important reason why birds sit on their eggs?
\par
\red{Here is an answer:}\par
to keep the eggs warm
\par
\red{Provide rationale for the above question and answer:}
\\\cline{1-2}

Proofwriter 
&   
Charlie is not furry. Charlie is smart. Charlie is not white. Charlie is not young. Dave is not white. Fiona is rough. Harry is not furry. All smart things are not cold. If Fiona is rough then Fiona is young. All furry, white things are smart. Young things are furry. Furry, young things are white. If something is young and kind then it is rough.

\red{Here is a hypothesis:} Fiona is furry.

\red{Is the hypothesis correct? Here are three options:}
No.
Yes.
Unknown.
\par
\red{Choose the correct option and justify your choice:}
\\\cline{1-2}

GSM8K 
&   
Jolene and Phil have four children, each with the same birthday.  They gave birth to their first child exactly 15 years ago.  They gave birth to their second child exactly one year after the birth of their first child.  They gave birth to their third child on the fourth birthday of their second child. Two years after the birth of their third child, they gave birth to their fourth child.  How old, in years, is their fourth child?
\\\cline{1-2}

PIQA 
&   How do I build walls when making a sunroom addition to a house?
\\\cline{1-2}

CosmosQA 
&   
This week has been a heck of a lot better than last week . Actually , it 's been the best week I ' ve had in ages . On Tuesday I went into town with Debbie , mostly charity shop shoping , bought pretty much a whole outfit . I wore the top / dress thing that I bought then on Wednesday when I went into London with Drew .

\red{Given the above context, }What is meant by charity shop shopping with Debbie ?
\\\cline{1-2}

e-SNLI 
&   
\textcolor{red}{Here is a premise:}\par
Two dogs, one carrying a large dish in its mouth.

\red{Here is a hypothesis:}\par
A dog carrying a large dish and another dog.

\red{Does this premise imply the hypothesis? Please justify your answer:}
\\\cline{1-2}

Adversarial NLI 
&   
\textcolor{red}{Here is a premise:} \par
Rafael Dumas (born 13 March 1995) is a Brazilian professional footballer who plays as a defender for Paysandu, on loan from Brazilian Série A club Flamengo. He previously had a six-month loan spell at FC Goa in the Indian Super League in 2016 and at Global.

\red{Here is a hypothesis:}\par
Dumas lived in Brazil for six months before moving to India.

\red{Does this premise imply the hypothesis? Please justify your answer:}
\\\cline{1-2}

ECQA 
& 
John put the book in his pack and then went off to work.  Where might he be working?

\red{Here are the options:}\par
Option 1: public library\par
Option 2: backpack\par
Option 3: briefcase\par
Option 4: suitcase\par
Option 5: synagogue\par

\textcolor{red}{Please choose the correct option and justify your choice:}
\\\cline{1-2}

GPT-3 summarization 
&   \textcolor{red}{Give a summary of the below article:}\par
A man is being sought by Kent Police after a house fire in which a person died.  Kent Fire and Rescue Service were called to the address in Shorncliffe Road, Folkestone, just before 10:30 GMT on Thursday. The victim was found inside the property after firefighters put out the blaze. Police said they believed Liam Petts, 24, may have information about the fire. Kent Police said the cause of the fire is not yet known, but is being treated as suspicious.
\\\cline{1-2}

DeFacto 
&   \textcolor{red}{Give a summary of the below article:}\par
Up to eight people are believed to have carried out the ram-raid at Martins newsagent in Prentice Place, Harlow, at about 01:45 BST. The ATM was removed from an outside wall, causing ``substantial damage'', and was then loaded on to a van. Essex Police are investigating whether the raid is linked to another at a Roydon newsagent earlier this month. The cash machine contained a ""five-figure sum"", police said. It is thought the thieves left in the van and a red car, believed to be an Audi. The digger, which was damaged in the raid, was left at the scene. ``We are especially keen to hear from anyone who saw the JCB, which is not the usual sort of vehicle to be seen on public roads late at night,'' Det Insp Jim Adams said. ``There was a previous incident involving the theft of a cash machine at a newsagent in Roydon on 16 September in which the machine and its contents were later recovered, and we are looking into whether these two incidents may be linked.''
\\

\bottomrule
\end{tabular}
}
\end{center}
\caption{Example prompts from datasets that are used in the human annotation. Templates of prompting LLMs to obtain the candidate outputs are in red. Note that for PIQA, CosmosQA, GPT-3 summarization, and DeFacto there is no need to prompt LLMs to obtain candidate outputs because those datasets already contain both correct and wrong answers. The templates are included as ease of contexts for human annotators. 
}
  \label{tab:human_data_dist}%
\end{table*}

\begin{figure*}[ht]
\begin{center}
\scalebox{0.90}
	    {
\fbox{\begin{minipage}[b]{\linewidth}

\paragraph{Goal}
We would like to collect natural language feedback for a diverse set of contexts and outputs. Given a context, we will ask annotators to provide natural language feedback about errors in the output answers. 
\newline

\paragraph{Guidelines}
For each question we provide a context, a correct output, and a candidate output. We are asking you to give feedback on whether there are any errors in the candidate output. The correct output is provided to help you identify potential errors in the candidate output more easily. If there are no errors in the candidate output, please also provide your feedback that there are no errors. Remember your main goal is to provide feedback so that a model can use your feedback to improve its generation. Please ignore the minor mistakes (e.g. missing punctuations, etc) and focus on the critical ones.
\newline

\paragraph{Annotation task}
\begin{enumerate}[topsep=0pt,noitemsep] 
    \item Choose the error types and write your feedback based on the chosen error types.
    \item Please flag the following issues if there's any. 
    \begin{enumerate}[topsep=0pt,noitemsep]
        \item The context is too complex to work on.
        \item Inappropriate content.
        \item Candidate output not understandable at all.
        \item Errors in the correct output.
    \end{enumerate}
\end{enumerate}

\paragraph{To give a good critique, please}
\begin{enumerate}[topsep=0pt,noitemsep] 
    \item Select from the provided error types which ones appear in the candidate output.
    \item Describe where the error(s) occur specifically, quote the generated content if necessary. Be detailed enough that someone else could easily find it.
    \item Justify your reason. Please be specific and explain why it is wrong.
    \item To the best of your knowledge, describe how to correct the error. Please flag the example if it's too complex for you. 
    \item When writing the critique, please only focus on the candidate output; please do not copy/paste the correct output.
    \item To the best of your knowledge, identify all the errors in the candidate output. It is okay to skip minor issues but you need to identify critical ones. Critical errors are ones that by fixing them, we can get a mostly correct answer. 
\end{enumerate}

\paragraph{To avoid a bad critique, please}
\begin{enumerate}[topsep=0pt,noitemsep] 
    \item Don’t just copy and paste the correct output.
    \item Don’t just provide corrections without justification.
    \item Don’t be vague, include references to the errors if necessary. 
\end{enumerate}

\end{minipage}}
}
\caption{The overall annotation instruction.}
\label{fig:guide}
\end{center}
\end{figure*}

\begin{table*}[p]
\begin{center}
\renewcommand{\arraystretch}{1.2}
\scalebox{0.85}
	    {
\begin{tabular}{@{}p{120pt}p{320pt}}
\toprule
{Error Type}
& {Description} \ \
\\ \midrule

Arithmetic 
& Error in math calculations.
\\\cline{1-2}

Coherence and deduction
& Sentences that do not logically follow each other, a summary that lacks a clear topic or conclusion, no structure, steps contradict, etc. This also includes \textit{Missing Step} that a step in a reasoning/explanation or thought process is missing (typically observed in math or logical reasoning problems).
\\\cline{1-2}

Consistency with context
& Information about an object (i.e., quantity, characteristics) or a personal named entity does not match information provided in the context/question.
\\\cline{1-2}

Veracity
& Information is not provided in the context and is irrelevant or wrong. 
For our annotation task rather than needing to look up, please just refer to the correct output which we assume to be the gold answer.
\\\cline{1-2}

Redundancy
& Explanation contains redundant information, which even though may be factual, is not required to answer the question and/or repeated in the output.
\\\cline{1-2}

Commonsense
& The output lacks relations that should be known from the general world. Should be instinctive, without questioning it, based on belief, and accepted by the society, e.g. all ducks are birds.
\\\cline{1-2}

No error
& The output is correct.
\\

\bottomrule
\end{tabular}
}
\end{center}
\caption{
Error types for human data collection. 
}
  \label{tab:error_types}%
\end{table*}

\begin{table*}[p]
\begin{center}
\renewcommand{\arraystretch}{1.2}
\scalebox{0.85}
	    {
\begin{tabular}{@{}p{120pt}p{120pt}p{170pt}@{}}
\toprule
{Dataset}
& {\# of prompts collected} \ \
& {Error types distribution} \ \  
\\ \midrule

Entailment Bank  
& 11
& Coherence and Deduction (55\%)\par
Commonsense (9\%)\par
Veracity (36\%)
\\\cline{1-3}

Proofwriter 
& 162
& Coherence and Deduction (88\%)\par
Commonsense (1\%)\par
Veracity (11\%)
\\\cline{1-3}

GSM8K 
& 431
& Arithmetic (33\%)\par
Coherence and Deduction (63\%)\par
Commonsense (3\%)\par
Veracity (1\%)
\\\cline{1-3}

PIQA 
& 246
& Coherence and Deduction (44\%)\par
Commonsense (41\%)\par
Veracity (15\%)
\\\cline{1-3}

CosmosQA 
& 143
& Coherence and Deduction (64\%)\par
Commonsense (21\%)\par
Veracity (15\%)
\\\cline{1-3}

e-SNLI 
& 65
& Coherence and Deduction (57\%) \par
Commonsense (33\%) \par
Veracity (10\%)
\\\cline{1-3}

Adversarial NLI 
& 68
& Arithmetic (1\%) \par
Coherence and Deduction (61\%) \par
Commonsense (24\%) \par
Veracity (14\%)
\\\cline{1-3}

ECQA 
& 118
& Coherence and Deduction (73\%) \par
Commonsense (17\%) \par
Veracity (10\%)
\\\cline{1-3}

GPT-3 summarization 
& 26
& Coherence and Deduction (69\%) \par
Veracity (31\%)
\\\cline{1-3}

DeFacto 
& 29
& Coherence and Deduction (52\%) \par
Veracity (48\%)
\\

\bottomrule
\end{tabular}
}
\end{center}
\caption{
Distribution of collected prompts and the identified error types from each dataset used in human annotation. 
}
  \label{tab:number_errortypes}%
\end{table*}

\begin{table*}[p]
\begin{center}
\renewcommand{\arraystretch}{1.2}
\scalebox{0.58}
	    {
\begin{tabular}{@{}p{55pt}p{360pt}p{180pt}p{150pt}@{}}
\toprule
{Error Type}
& {Description} \ \
& {Example good critique} \ \
& {Example bad critique} \ \ 
\\ \midrule

Arithmetic 
& \textbf{Context}: Natalia sold clips to 48 of her friends in April, and then she sold half as many clips in May. How many clips did Natalia sell altogether in April and May? \par
\textbf{Output (correct)}: 
Natalia sold 48/2 = 24 clips in May.
Natalia sold 48+24 = 72 clips altogether in April and May.
Answer: 72
\par
\textbf{Output (candidate)}: 
Natalia sold 48/2 = 22 clips in May.
Natalia sold 48+22 = 70 clips altogether in April and May.
Answer: 70
& Operation in the first step is incorrect. 48/2 equals 24. Hence the final answer should be 48+24=72.  
& 48/2=24 \par
\par \hspace{1cm} \par
\textcolor{red}{This critique is bad because there is only correction but not justification.}
\\\cline{1-4}

Coherence and deduction
& \textit{Example 1 (lack of logic)} \par
\textbf{Context}: Here is a dialogue:\par
W: I am considering dropping my dancing class. I am not making any progress.\par
M: If I were you, I would stick with it. It's definitely worth time and effort.\par

What does the man suggest the woman do?\par

\textbf{Output (correct)}: 
Continue her dancing class.
\par
\textbf{Output (candidate)}: 
Take a more interesting class.
\par \hspace{1cm} \par
\textit{Example 2 (missing steps)}\par
\textbf{Context}: \par
Anne is furry. Anne is green. Anne is quiet. Charlie is green. Charlie is quiet. Fiona is furry. Harry is white. If someone is nice then they are white. All nice, rough people are round. If Fiona is quiet then Fiona is green. If someone is quiet and round then they are nice. All rough people are quiet. If someone is furry then they are rough. 
\par
Here is a hypothesis: Anne is rough.
\par
Is the hypothesis correct?
\par
\textbf{Output (correct)}: 
The answer is Yes because Anne is furry. If someone is furry then they are rough. Anne is rough.
\par
\textbf{Output (candidate)}: 
The answer is Yes because Anne is rough.
& \textit{Example 1} \par
The candidate output does not follow the context logically. The woman mentions the dancing class and the man suggests to the woman to stick with it, so it is not implied that she should quit dancing and start a different class.
\par \hspace{1cm} \par
\textit{Example 2} \par
Missing steps when determining whether Anne is rough. Anne is rough according to ``Anne is furry.'' and ``If someone is furry then they are rough.''

& \textit{Example 1}\par
Continue her dancing class. \par
\par \hspace{1cm} \par
\textcolor{red}{This critique is bad because it just copies the correct output.}

\par \hspace{10cm} \par
\par \hspace{10cm} \par
\par \hspace{3cm} \par

\textit{Example 2}\par
Missing steps. \par
\par \hspace{1cm} \par
\textcolor{red}{This critique is bad because it just points out the error type but doesn't mention where the error is.}
\\\cline{1-4}

Consistency with context
& \textbf{Context}:
Locsin and Anakin went fishing. Locsin caught 5 fish and Anakin caught 10 fish. How many fish did Anakin catch?

\textbf{Output (correct)}: 
Anakin caught 10 fish.

\textbf{Output (candidate)}: 
Anakin caught 5 fish.
& The candidate output contradicts with the information mentioned in the context that Locsin caught 5 fish and Anakin caught 10 fish. So Anakin caught 10 fish.
& Anakin caught 10 fish.
\par \hspace{1cm} \par
\textcolor{red}{This critique is bad because there is only correction but no justification.}
\\\cline{1-4}

Veracity
& \textbf{Context}:
The basketball team went to the steakhouse to eat dinner. The first player ate a 6-ounce steak. The second player ate beef tips, containing 8 beef tips, each an ounce in size. The third player ate a one-pound steak. And the fourth and fifth players ordered vegetarian meals. In total, how many ounces of meat were consumed by the team?

\textbf{Output (correct)}: 
The first player ate 6 ounces of steak. 
The second player ate 8 ounces of beef tips, with each tip being an ounce, for a total of 8 x 1 = 8 ounces.
The third player ate a 16-ounce steak, which is equivalent to 16 x 1 = 16 ounces.
Therefore, the total amount of meat consumed by the team is:
6 + 8 + 16 = 30 ounces.
The fourth and fifth players ordered vegetarian meals, so they did not consume any meat.

\textbf{Output (candidate)}: 
The fourth and fifth players ordered vegetarian meals, for a total of 2 ounces of meat.
& The candidate output does not answer the question of “how many ounces of meat were consumed by the team”, but instead makes up facts that the fourth and fifth players ordered 2 ounces of vegetarian meals, which is irrelevant and also not true because the amount of vegetarian meals is not mentioned in the context. It should calculate the total amount of meat by calculating the amounts that the first three players consumed. 
& The fourth and fifth players did not order 2 ounces of meat. 
\par \hspace{1cm} \par
\textcolor{red}{This critique is bad because it does not give a correction to the error made in the candidate output and also it does not specify the error types.}  
\\\cline{1-4}

Redundancy
& \textit{Example 1} (redundant) \par
\textbf{Context}:
Bob is cold. If something is quiet then it is blue.
All blue things are round. If something is round and smart then it is cold. 

Question: is Bob cold?

\textbf{Output (correct)}: 
The answer is true because Bob is cold, as mentioned in the context.

\textbf{Output (candidate)}: 
If something is round and not smart then it is blue. All blue things are round. Answer: true.

\textit{Example 2} (repetition) \par
\textbf{Context}: Parker chews 4 pieces of gum a day. A pack of gum has 15 pieces of chewing gum per pack. How many packs of gum will he need to last him 30 days?

\textbf{Output (correct)}: 
If Parker chews 4 pieces of gum per day, he will need 120 pieces of gum to last him 30 days. To determine the number of packs of gum he needs, we divide the total number of pieces needed by the number of pieces per pack, which is 15. The calculation would be 120 $\div$ 15 = 8 packs of gum. Therefore, Parker needs to buy 8 packs of gum to last him 30 days if he chews 4 pieces of gum per day.

\textbf{Output (candidate)}: 
He chews 4 pieces of gum a day and there are 30 days in a month so he chews 4 $\times$ 30 = 120 pieces of gum a month. Parker chews 120 pieces of gum a month. He needs 120 pieces of gum a month and a pack of gum has 15 pieces of gum so he will need 120/15 = 8 packs of gum.
& \textit{Example 1}\par
The reasoning in the candidate output is irrelevant to the question about whether Bob is cold or not. The candidate output should not mention blue and round things but should answer the question directly, which was already provided in the context.
\par \hspace{1cm} \par
\textit{Example 2}\par
There is repetition in the candidate output. Parker chewing 120 pieces of gum a month is mentioned in the first and second sentences.
& \textit{Example 1}\par
Why mention blue and round things when the question asks about whether Bob is cold?
\par \hspace{1cm} \par
\textcolor{red}{This critique is bad because it asks a question rather than giving a critique and correction.}

\par \hspace{1cm} \par
\textit{Example 2}\par
The candidate output repeats itself.
\par \hspace{1cm} \par
\textcolor{red}{This critique is bad because it does not point out where the error is.}  
\\\cline{1-4}

Commonsense
& \textbf{Context}:
How do you taste something?

\textbf{Output (correct)}: 
Place it in your mouth to taste.

\textbf{Output (candidate)}: 
Smell it enough to taste it.
& The candidate output contradicts with common sense that to taste something you should place it in your month rather than smell it.
& How do you taste something if you smell it?
\par \hspace{1cm} \par
\textcolor{red}{This critique is bad because there is no justification.}  
\\\cline{1-4}

No error
& \textbf{Context}:
Bob is big. Bob is smart. Dave is cold. Dave is round. Dave is smart. Dave is young. Erin is cold. Erin is round. Erin is smart. Harry is green. If someone is round then they are white. Big people are cold. Big, round people are cold. All cold people are round. Smart, white people are green. If Harry is green then Harry is cold. All round people are white. All green, smart people are young.

Here is a hypothesis: Dave is big.

Is the hypothesis correct?

\textbf{Output (correct)}: 
The answer is Unknown because There is not enough evidence to provide an answer for this question given the context. So the answer is unknown.

\textbf{Output (candidate)}: 
There is not enough context to determine whether Dave is big. The answer is unknown.
& There is no error in the candidate output.
& (empty) 
\par \hspace{1cm} \par
\textcolor{red}{This critique is bad because there is no critique. Annotators should provide critique even if there is no error in the candidate output.}  
\\

\bottomrule
\end{tabular}
}
\end{center}
\caption{
Guidelines with examples provided to annotators. Annotators are asked to provide critique on the candidate output. To ensure better quality of data collection, we provide a golden correct output to help annotators identify errors in the candidate output. Both good and bad examples of critiques are provided. The issue of the example bad critiques are highlighted in red. 
}
  \label{tab:human_examples}%
\end{table*}

\section{Evaluation Pipeline}
\subsection{Instruction Derivation}
\label{sec: instruction_derivation}
We show a few examples in Table~\ref{tab:instruction_examples}. With the instruction manually tuned by us, GPT-4 can give reasonable scores on these examples.
We provide examples where GPT-4 demonstrates the limitations of model-based evaluation in Table~\ref{tab:gpt4_failure} and Table~\ref{tab:gpt4_failure_2}.

\begin{table*}[ht]
    \centering
    \scriptsize
    \begin{tabular}{l}
    \toprule
    \#\#\# Instruction: Edit the following sentence so that it expresses an opinion: \\
    \#\#\# Input:Social media can be both positive and negative. \\
    \#\#\# Response:\\
    \#\#\# Feedback:I think social media can be both a blessing and a curse. It has the power to connect people, \\spread information, and allow us to express ourselves, but it can also be used to bully, spread misinformation, and distort people's self-images.\\
    GPT-4 score: 3\\
    \midrule
     \#\#\# Question: What are some great financial investments with no risk at all?\\
     \#\#\# Response: Investing in the stock market, mutual funds, bonds, and real estate are all great financial investments with no risk at all. \\
    \#\#\# Feedback:The response is not entirely accurate. Investing in the stock market, mutual funds, bonds, \\and real estate do have some level of risk. The response should be revised to reflect this.\\
    GPT-4 score: 7\\
    \midrule
    \#\#\# Instruction: Please share your feedback regarding the following question and answer.\\
    \#\#\# Question: What are some great financial investments with no risk at all?\\
    \#\#\# Response: Investing in the stock market, mutual funds, bonds, and real estate are all great financial investments with no risk at all.\\
    \#\#\# Feedback:The response provides a general answer to the question but lacks specific examples or details. \\It would be helpful to provide more diversified investment options and explain the potential benefits and risks of each.\\
    GPT-4 score: 2\\
    \midrule
    \#\#\# Instruction: Identify the right order of reading for the given set of books. \\
    \#\#\# Input: Lolita, Great Gatsby, Catcher in the Rye \\ 
    \#\#\# Response:The right order of reading is Great Gatsby, Catcher in the Rye, Lolita.\\
    \#\#\# Feedback:The right order of reading is Great Gatsby, Catcher in the Rye, Lolita.\\
    GPT-4 score: 1\\
    \midrule
    \#\#\# Question: What are the differences between plant-based and animal-based protein sources? \\
    \#\#\# Answer: The main difference between plant-based and animal-based protein sources is that plant-based proteins are derived from plants, \\while animal-based proteins are derived from animals. Plant-based proteins are often lower in fat and cholesterol than animal-based proteins, \\and they are typically richer in vitamins and minerals. Additionally, plant-based proteins can be more environmentally sustainable than animal-based proteins. \\
    \#\#\# Feedback: The response doesn't mention the most important difference which is that animal-based protein sources are complete protein sources, \\and plant-based protein sources are incomplete.\\
    GPT-4 score: 7\\
    \midrule
    \#\#\# Instruction: Rewrite the first sentence to make it active voice. \\
    \#\#\# Input: The drapes were brown in color.\\
    \#\#\# Answer:I chose the browns drapes to add a touch of warmth to the room.\\
    \#\#\# Feedback:The proposed answer is clear and concise, but it does not address the prompt to rewrite the sentence in active voice. A more appropriate response would be: \\``I chose brown drapes to add a touch of warmth to the room.'' This sentence is in active voice and effectively conveys the same information as the original sentence.\\
    GPT-4 score: 6\\
    \bottomrule
    \end{tabular}
    \caption{Examples of likert scale scores given by GPT-4. }
    \label{tab:instruction_examples}
\end{table*}

{\small
\begin{table*}[]
\centering
\scalebox{0.7}{
\begin{tabular}{|p{1.5cm}|p{8cm}|}\hline
Question               & A football team played 22 games. They won 8 more than they lost. How many did they win? \\ \hline
Response               & 22.                      \\ \hline
Feedback               & The answer is correct, but it could benefit from a brief explanation. To find out how many games the team won, we can subtract the number of losses from the total number of games. In this case, since the team won 8 more games than they lost, we can subtract 8 from the total number of games played (22). Therefore, the team won 14 games.\\ \hline
Ground truth    & 15. \\ \hline
GPT4 Score                  & 5: The feedback misses the important errors but clearly confirms that the answer is wrong.                          \\ \hline
\end{tabular}}
\caption{GPT-4 evaluation failure example. Even though the ground truth is provided, GPT-4 still assigns a high score to the incorrect feedback.}
\label{tab:gpt4_failure}
\end{table*}
}

{\small
\begin{table*}[]
\centering
\scalebox{0.7}{
\begin{tabular}{|p{4cm}|p{18cm}|}
 \hline
Question               & Are there any papers where two (or more) authors share the same full name? \\ \hline
Response               & 
\begin{minipage}[t]{\textwidth}
Here are few examples:
\begin{itemize}[nosep]
\item \url{https://scholar.google.com/scholar?hl=en&as_sdt=0\%2C5&q=author\%3A\%22Frank+Frank\%22&btnG=} 420 authors named Frank Frank
\item \url{https://scholar.google.com/scholar?hl=en&as_sdt=0\%2C5&q=author\%3A\%22John+John\%22&btnG=} 2480 authors named John John
\item \url{https://scholar.google.com/scholar?hl=en&as_sdt=0\%2C5&q=author\%3A\%22William+William\%22&btnG=} 5960 authors named William William
\end{itemize}
\vspace{0.1cm}
\end{minipage} \\ \hline
SelFee Feedback & The provided answer is informative and gives a good starting point for further research. However, it would be helpful to provide more context or a specific example related to the question asked. Additionally, it would be beneficial to include information on how to handle situations where two or more authors have the same name in a citation or research paper.\\ \hline
ChatGPT Feedback       & The response provided is informative and relevant to the question, providing multiple examples of papers where authors share the same full name. The links provided make it easy for the reader to access the papers and verify the information. Overall, the answer is well-structured and helpful. \\ \hline
\ourmodel~ Feedback     & I'm not sure if the answers to my questions are right. I'm looking for papers where two (or more) authors share the same full name. I'm not looking for papers where two (or more) authors have the same first name, and where they have the same last name. \\ \hline
\end{tabular}}
\caption{With the above example, GPT-4 gives SelFee score 6, ChatGPT score 4, \ourmodel{} score 2. However, only \ourmodel{} successfully points out the error in the answer.}
\label{tab:gpt4_failure_2}
\end{table*}
}
\end{document}